\documentclass[runningheads]{llncs}
\usepackage{booktabs}
\usepackage{multirow}
\usepackage[T1]{fontenc}
\usepackage{graphicx,verbatim}
\usepackage{amsfonts}
\usepackage{amsmath} 

\usepackage{marvosym}
\usepackage{color}
\usepackage{hyperref}

\begin{document}
\title{ScribbleDose: Scribble-Guided Dose Prediction in Radiotherapy}
\author{
Zhenxi Zhang\inst{1}\thanks{Equal contribution.} \and
Yitao Zhuang\inst{1}$^{\star}$ \and
Yao Pu\inst{1}$^{\star}$ \and
Peixin Yu\inst{1}$^{\star}$ \and
Zirong Li\inst{3} \and
Yan Xia\inst{3} \and
Hui Li\inst{1} \and
Bin Li\inst{4} \and
Fuchen Zheng\inst{5} \and
Ge Ren\inst{1,2}\thanks{Corresponding author.}
}


\authorrunning{Zhang et al.}
%
\institute{
Department of Health Technology and Informatics, The Hong Kong Polytechnic University, Hong Kong SAR
\and
The Hong Kong Polytechnic University Shenzhen Research Institute, The Hong Kong Polytechnic University, China
\and
Department of Orthodontics and Orofacial Orthopedics, Friedrich-Alexander-University Erlangen-Nuremberg, Germany
\and
Institute of Scientific Instrumentation, Shenzhen Institutes of Advanced Technology, Chinese Academy of Sciences, China
\and
Department of Computer and Information Science, University of Macau, Macau SAR
}
  
\maketitle               

\begin{abstract}
Anatomical structure masks are widely adopted in radiotherapy dose prediction, as they provide explicit geometric constraints that facilitate structure–dose coupling. 
However, conventional manual delineation of these masks requires precise annotation of structure boundaries relevant to radiotherapy, which is time-consuming and labor-intensive.
To address these limitations, we propose a scribble-guided dose prediction framework that relies solely on anatomical structures annotated with sparse scribbles. 
Specifically, we design a Scribble Completion Module (SCM) to generate dense anatomical masks by propagating sparse scribble labels to semantically similar voxels.
During the propagation process, a supervoxel-based regularization is introduced to preserve geometric boundary consistency to ensure anatomical plausibility.
Furthermore, we propose a Structure-Guided Dose Generation Module (SGDGM) to strengthen the correspondence between sparse structural cues and dose distribution. 
Herein, the completed dense masks derived from scribbles serve as structural guidance to condition the dose prediction network.
This scribble–mask–dose consistency encourages high-dose concentration within target volumes while effectively sparing surrounding organs-at-risk.
Extensive experiments on the open-source GDP-HMM dataset demonstrate that the proposed method maintains superior dose prediction performance while substantially reducing annotation cost, providing a practical paradigm for dose prediction under sparse structural annotation.
The code and reannotated scribbles will be made publicly available upon the acceptance of this paper.

\keywords{Radiotherapy Dose Prediction \and Scribble Annotation \and Structure Guided Modeling.}

\end{abstract}
\section{Introduction}
Radiotherapy planning plays a central role in modern cancer treatment, as treatment outcomes largely depend on the delivered dose distribution~\cite{hernandez2020plan}.
A high-quality plan requires sufficient coverage of the Planning Target Volume (PTV) while minimizing irradiation to surrounding Organs-at-Risk (OAR)~\cite{monshouwer2010practical}. 
Traditional treatment planning relies on manual iterative optimization guided by expert-defined objectives, which is time-consuming and may lead to inter-planner variability~\cite{xiong2024automatic}. 
Recently, deep learning-based approaches built upon CNN~\cite{xiong2026generalizable,nguyen2019feasibility}, Transformers~\cite{hu2023trdosepred,wen2023transformer}, Mamba~\cite{zhou2025novel}, and diffusion model~\cite{zhang2024dosediff} have been widely adopted for automated dose prediction.
These models significantly shorten planning time while achieving accuracy comparable to manual plans, positioning deep learning-based dose prediction as a promising approach for radiotherapy planning~\cite{nijkamp2025current}.

\begin{figure}[t]
\includegraphics[width=\textwidth]{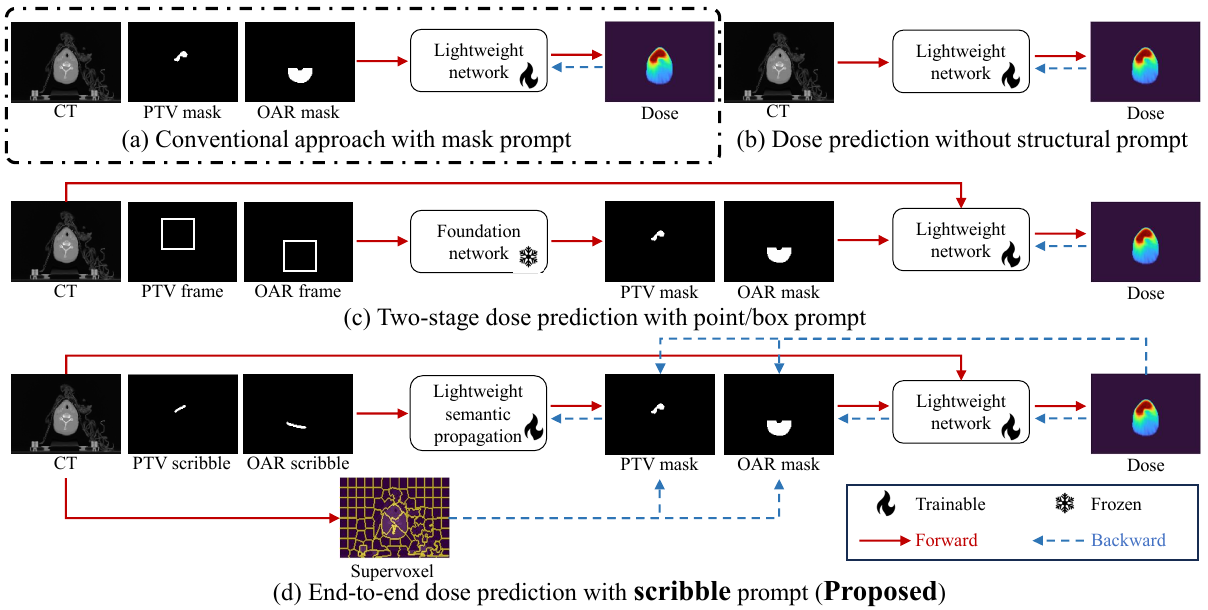}
\caption{Comparison of existing dose prediction paradigms and the proposed scribble-guided framework. 
(a) represents the conventional mask-based paradigm that relies on densely delineated structural masks.
(b) and (c) aim to reduce annotation burden through CT-only modeling and two-stage foundation-model-based pipelines with box prompts, respectively.
(d) presents the proposed unified scribble-guided framework.} \label{fig1}
\end{figure}
However, most existing deep learning-based dose prediction models rely on fully delineated anatomical structure masks as explicit geometric priors to model structure–dose relationships, as illustrated in Fig.~\ref{fig1}(a).
These masks provide critical spatial constraints that facilitate accurate dose allocation within target volumes and effective sparing of surrounding OARs.
Nevertheless, obtaining high-quality structural annotations requires precise manual delineation of radiotherapy-relevant boundaries, which is labor-intensive, time-consuming, and highly dependent on clinical expertise.
To alleviate this burden, Wllems et al. first proposed predicting dose distributions directly from CT images, as shown in Fig.~\ref{fig1}(b), without requiring explicit PTV or OAR masks~\cite{willems2019feasibility}. 
Subsequently, Jiao et al. introduced a CT-to-dose framework that incorporates supervoxel-based representations to enrich anatomical information while still avoiding manual contour inputs~\cite{jiao2023transdose}. 
While such mask-free approaches alleviate the need for manual delineation, the absence of explicit structural guidance may compromise the modeling of fine-grained geometric constraints that are critical for clinically plausible dose allocation.
To incorporate explicit structural constraints while reducing manual contouring, subsequent studies have adopted a two-stage framework based on foundation models~\cite{zhang2023segment,zhang2024generalizable}. 
In this framework, foundation models are first employed to extract dense PTV and OAR masks from CT images, often guided by sparse prompts such as points and boxes, as illustrated in Fig.~\ref{fig1}(c). 
The generated masks are subsequently used as inputs for downstream dose prediction.
Although this approach alleviates manual contouring during deployment, the two-stage design separates anatomical structure extraction from dose modeling, thereby weakening structure–dose coupling.

To address these limitations, we explore sparse scribble supervision as a lightweight alternative to fully delineated masks. 
Compared to dense mask annotations, scribble substantially reduces annotation cost by avoiding precise boundary delineation of PTVs and OARs while retaining essential radiotherapy-relevant structural cues~\cite{lin2016scribblesup,zhang2022cyclemix}. 
Given their sparse and coarse nature, scribble annotations provide limited boundary information, necessitating mechanisms to recover complete geometric representations for accurate dose modeling.
Building upon this insight, we design a scribble-guided dose prediction framework (Fig.~\ref{fig1}(d)) that progressively propagates sparse scribble cues into dense anatomical masks and establishes a continuous scribble-mask-dose optimization pipeline.
Specifically, the proposed Scribble Completion Module (SCM) propagates sparse scribble labels to semantically correlated voxels, generating dense structural masks.
To ensure anatomical plausibility, supervoxel-based boundary priors are derived from CT images to construct geometric constraints, guiding the propagated masks to align with inherent anatomical boundaries.
The completed masks are integrated into a Structure-Guided Dose Generation Module (SGDGM), which adopts an attention mechanism to incorporate structural guidance.
Precisely, attention responses are positively modulated by PTV masks and negatively modulated by OAR masks, promoting dose concentration in targets while limiting exposure to OARs.
Importantly, dose prediction losses are further propagated back to the scribble completion stage, enabling joint optimization across scribble-mask-dose modeling.
Extensive experiments on the benchmark dataset validate the effectiveness of the proposed scribble-guided framework, demonstrating that high-quality dose prediction can be achieved under sparse structural annotation.
Our contributions are summarized as follows:
\begin{enumerate}
    \item We reannotate and publicly release \textbf{scribble-level structural labels} on the GDP-HMM dataset, and propose a unified \textbf{scribble-guided dose prediction framework} that alleviates the need for fully delineated structural masks.
    \item We introduce a \textbf{Scribble Completion Module} that reconstructs anatomically consistent dense masks from sparse scribbles with supervoxel boundary constraints.
    \item We develop a \textbf{Structure-Guided Dose Generation Module} that utilizes mask-guided attention to couple reconstructed masks with dose, enabling continuous scribble–mask–dose learning.
\end{enumerate}

\begin{figure}[t]
\includegraphics[width=\textwidth]{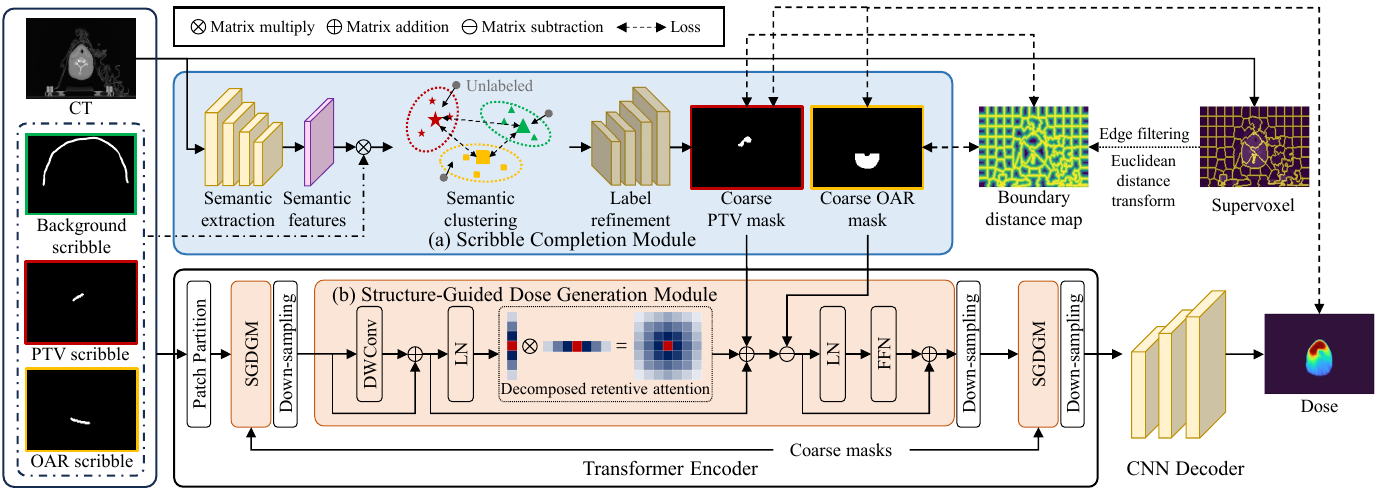}
\caption{Overview of the proposed scribble-guided dose prediction framework.
The framework includes two main components: (a) the Scribble Completion Module (SCM) and (b) the Structure-Guided Dose Generation Module (SGDGM).} \label{fig2}
\end{figure}
\section{Method}
In this section, we present the proposed scribble-guided dose prediction framework (Fig.~\ref{fig2}), which reformulates structure-aware dose modeling under sparse structural supervision. 
Rather than relying on fully delineated masks, the proposed framework establishes a continuous scribble–mask–dose learning pipeline that incorporates geometric constraints under sparse annotation and jointly optimizes structural reconstruction and dose generation.
The framework is built around two principal modules.  
The Scribble Completion Module (SCM, Fig.~\ref{fig2}(a)) reconstructs dense structural masks from sparse scribbles under supervoxel-based geometric constraints. 
The Structure-Guided Dose Generation Module (SGDGM, Fig.~\ref{fig2}(b)) then integrates the reconstructed masks into dose modeling via mask-aware attention. 
This design allows structural cues and dose supervision to mutually reinforce each other during training.
\subsection{Scribble Completion Module}
The Scribble Completion Module aims to reconstruct dense structural masks from sparse scribble annotations. 
To achieve this, SCM consists of two complementary components: (1) a scribble-guided semantic clustering strategy to propagate sparse labels in feature space, and (2) a supervoxel-based boundary constraint to enforce anatomical consistency in spatial domain.

\subsubsection{Scribble-Guided Semantic Clustering} aims to propagate sparse scribble annotations to unlabeled voxels in feature space and generate dense structural masks. 
Specifically, given a CT volume $X \in \mathbb{R}^{D \times H \times W}$, we extract volumetric features $F \in \mathbb{R}^{C \times D \times H \times W}$ utilizing a lightweight 3D encoder, where $C$ is the feature dimension.
For each semantic class $k \in \{\text{PTV}, \text{OARs}, \text{Background}\}$, we compute a class-wise feature centroid from scribble-labeled voxels. 
Let $\Omega_k$ denote the voxel set annotated by scribbles of class $k$, and $N_k$ its cardinality.
The centroid is defined as:
\begin{equation}
c_k = \frac{1}{N_k} \sum_{i \in \Omega_k} \frac{F_i}{\|F_i\|},
\end{equation}
where feature vectors are $\ell_2$-normalized to stabilize similarity computation.
We then measure the similarity between each voxel feature and the class centroids using cosine similarity:
\begin{equation}
s_{k,i} = \frac{c_k^\top F_i}{\tau},
\end{equation}
where $\tau$ is a temperature parameter. 
The similarity scores $s_{k,i}$ are computed for unlabeled voxels and serve as coarse class logits. 
These logits are subsequently refined by a lightweight decoder to obtain dense mask predictions.

To encourage intra-class compactness and inter-class separability in feature space, we introduce two regularization terms. 
The compactness loss reduces feature variance within scribble regions:
\begin{equation}
\mathcal{L}_{\text{compact}} = 
\sum_k \frac{1}{N_k} \sum_{i \in \Omega_k} \|F_i - c_k\|^2,
\end{equation}
while the separation loss penalizes similarity between different class centroids:
\begin{equation}
\mathcal{L}_{\text{sep}} = 
\sum_{k \neq j} \exp\!\left(-\|c_k - c_j\|^2\right).
\end{equation}
Through this centroid-driven semantic propagation, sparse scribble annotations are progressively expanded into coarse structural masks in feature space.

\subsubsection{Supervoxel-Based Boundary Constraint} introduces geometry-aware spatial regularization into mask reconstruction. 
In particular, supervoxels partition the CT volume into locally homogeneous intensity regions, preserving intrinsic anatomical boundaries. 
By leveraging supervoxel-derived boundary cues, the reconstructed masks are encouraged to align with anatomically consistent spatial structures.

Direct alignment with hard boundary maps treats all non-boundary deviations uniformly, without distinguishing between small and large spatial discrepancies.
To provide distance-sensitive geometric supervision, we construct a distance-based boundary representation derived from supervoxel boundaries.
Specifically, we apply Euclidean Distance Transform (EDT) to the supervoxel boundary map to compute a distance field $D_{\text{sv}}$, where each voxel encodes its minimal distance to the nearest supervoxel boundary. 
The resulting distance map satisfies the Eikonal property $|\nabla D_{\text{sv}}| = 1$ almost everywhere, yielding a spatially continuous geometric supervision signal.
In practice, Gaussian smoothing is employed as an efficient approximation of distance-based weighting to obtain a differentiable soft boundary representation $\tilde{D}_{\text{sv}}$. 
The boundary constraint is formulated as:
\begin{equation}
\mathcal{L}_{\text{boundary}} =
\left\| B_{\text{pred}} - \tilde{D}_{\text{sv}} \right\|
\end{equation}
where $B_{\text{pred}}$ denotes the predicted boundary map. 
This distance-based regularization stabilizes boundary refinement and promotes anatomically plausible mask reconstruction beyond scribble-labeled regions.

\subsection{Structure-Guided Dose Generation Module}
The Structure-Guided Dose Generation Module aims to explicitly couple anatomical structures with dose modeling while preserving spatial continuity of dose distributions. 
To model such spatial coherence, SGDGM is implemented using Retentive Transformers (RMT)~\cite{fan2024rmt}. 
Specifically, RMT adopts decomposed retentive attention with distance-aware decay along spatial directions. 
This distance-decay mechanism encourages each patch to focus on neighboring regions while attenuating distant interactions, introducing an inductive bias aligned with the smoothness of dose distributions.

Building upon the RMT backbone, we introduce a structure-guided attention mechanism based on the reconstructed coarse masks.
Specidically, let $M_{\text{PTV}}$ and $M_{\text{OAR}}$ denote the coarse masks of PTV and OAR regions, respectively. 
The structure-guided modulation map $G$ is defined as:
\begin{equation}
G = 2 \cdot \sigma\!\left(1 + \alpha M_{\text{PTV}} - \beta M_{\text{OAR}}\right),
\end{equation}
where $\alpha$ and $\beta$ are learnable positive coefficients and $\sigma(\cdot)$ denotes the sigmoid function. 
The modulation map is applied multiplicatively to the multi-scale retentive attention maps:
\begin{equation}
\text{Attn}'_l = \text{Attn}_l \odot G_l,
\end{equation}
where $G_l$ is resized to match the spatial resolution of $\text{Attn}_l$. 
This modulation enhances attention responses within PTV regions while suppressing responses in OAR regions, thereby establishing explicit structure–dose coupling.

Moreover, to further reinforce structural consistency in dose allocation, we introduce a ranking-based regularization term:
\begin{equation}
\mathcal{L}_{\text{rank}} = \max\left(0, \bar{D}_{\text{OAR}} - \bar{D}_{\text{PTV}}\right),
\end{equation}
where $\bar{D}_{\text{PTV}}$ and $\bar{D}_{\text{OAR}}$ denote the average predicted dose within PTV and OAR regions, respectively. 
This auxiliary constraint encourages higher dose concentration in target volumes while suppressing excessive dose in sensitive structures.

\section{Experiments}
\subsection{Dataset and Implementation}
Experiments were conducted on the dataset released by the GDP-HMM 2025 challenge~\cite{gao2025automating}, which contains 3231 radiotherapy plans. 
Following the official partition, 2627 cases were used for training, 148 for validation, and 356 for testing. 
For each case, the dataset provides CT volumes together with PTV and OAR masks, as well as the corresponding clinical dose distributions.
Additionally, scribbles were generated slice-wise from manual masks following~\cite{boettcher2024scribbles}, ensuring they remain strictly inside the original contours.
Model training was conducted using PyTorch 2.6 on four NVIDIA A6000 GPUs. Optimization was performed with Adam for 300 epochs. The learning rate was initialized at $1\times 10^{-4}$ and gradually annealed to $1\times 10^{-6}$.
\subsection{Experimental Results}
\begin{table}[t]
\centering
\scriptsize
\setlength{\tabcolsep}{1.2pt} 
\caption{Performance comparison under different structure prompt types on the GDP-HMM dataset. 
* denotes a two-stage framework where segmentation is first performed to generate dense structural masks, followed by MedNeXt for dose prediction.
Values are reported as mean (standard deviation).}
\begin{tabular}{ccccccc}
\toprule
Method & Prompt & Dose Score~$\downarrow$ & DVH Score~$\downarrow$ & HI~$\downarrow$ & CI~$\downarrow$ & $D_{95}~\downarrow$ \\
\midrule

3D UNet~\cite{willems2019feasibility}      & \multirow{3}{*}{None} 
& 3.275(1.295) & 5.778(2.070) & 0.082(0.049) & 0.340(0.185) & 5.136(3.053) \\

RMT~\cite{fan2024rmt}  &  
& 3.516(1.368) & 7.085(3.215) & 0.157(0.158) & 0.526(0.223) & 6.709(7.324) \\

MedNeXt~\cite{roy2023mednext}        &  
& 3.152(1.651) & 6.425(2.294) & 0.100(0.076) & 0.393(0.185) & 5.142(3.662) \\

\midrule

MedSAM2*\cite{MedSAM2} & Box
& 2.670(1.356) & 5.382(2.099) & 0.157(0.064) & 0.587(0.236) & 7.537(4.521) \\

SAMMed3D*\cite{wang2025sam} & Point
& 2.732(1.275) & 5.219(1.784) & 0.143(0.061) & 0.576(0.248) & 6.558(3.957) \\

\midrule

C3D~\cite{liu2021cascade} & Mask
& 1.993(1.139) & 2.642(1.766) & \textbf{0.040(0.091)} & 0.557(0.285) & 2.398(1.980) \\

\midrule

C3D~\cite{liu2021cascade}           & \multirow{4}{*}{Scribble}
& 2.348(1.232) & 6.716(2.280) & 0.274(0.186) & 0.361(0.253) & 4.332(3.994) \\

RMT~\cite{fan2024rmt} & 
& 2.557(1.263) & 4.600(2.504) & 0.109(0.073) & 0.515(0.288) & 6.138(4.914) \\

MedNeXt~\cite{roy2023mednext}      & 
& 2.268(1.245) & 4.082(2.331) & 0.100(0.076) & 0.434(0.247) & 5.675(4.517) \\

ScribbleDose  & 
& \textbf{1.974(1.149)} & \textbf{2.471(1.641)} & $\underline{0.046(0.043)}$ & \textbf{0.171(0.188)} & \textbf{2.350(2.137)} \\

\bottomrule
\label{tab1}
\end{tabular}
\end{table}

We evaluate different structure prompt types using Dose score, Dose Volume Histogram (DVH) score, Homogeneity Index (HI), Conformity Index (CI), and $D_{95}$. 
Here, $D_{95}$ represents the dose received by 95$\%$ of the volume of PTV.
As shown in Table~\ref{tab1}, refining coarse prompts with foundation models in a two-stage pipeline noticeably improves performance. 
Specifically, MedSAM2 and SAMMed3D reduce the Dose score of MedNeXt without structural prompts from 3.152 to 2.670 and 2.732, respectively, indicating that transforming weak prompts into dense masks provides more effective geometric constraints for dose estimation.
Furthermore, the scribble-based unified framework achieves consistently lower Dose and DVH scores compared with the two-stage approach. 
In particular, ScribbleDose attains the best overall performance (Dose Score: 1.974, DVH Score: 2.471), outperforming all coarse-prompt methods. 
More importantly, even with sparse scribble prompts, ScribbleDose surpasses the mask-based C3D model, demonstrating that the proposed framework achieves superior performance with substantially reduced annotation requirements.
\begin{table}[t]
\centering
\scriptsize
\setlength{\tabcolsep}{3pt}
\caption{Ablation study on the Semantic Guidance (SG) and  Morphological Guidance (MG). Values are reported as mean (standard deviation).}
\begin{tabular}{ccccc ccc}
\toprule
SG & MG & Dose Score~$\downarrow$ & DVH Score~$\downarrow$ & HI~$\downarrow$ & CI~$\downarrow$ & $D_{95}~\downarrow$ \\
\midrule
$\times$ & $\times$ & 2.557(1.263) & 4.600(2.504) & 0.109(0.073) & 0.515(0.288) & 6.138(4.914) \\
$\checkmark$ & $\times$ & 2.209(1.202) & 3.329(2.049) & 0.078(0.057) & 0.268(0.228) & 3.913(3.476) \\
$\times$ & $\checkmark$ & 2.287(1.299) & 4.774(1.626) & 0.051(0.041) & 0.487(0.209) & 5.212(2.300) \\
$\checkmark$ & $\checkmark$ & \textbf{1.974(1.149)} & \textbf{2.471(1.641)} & \textbf{0.046(0.043)} & \textbf{0.171(0.188)} & \textbf{2.350(2.137)} \\
\bottomrule
\label{tab2}
\end{tabular}
\end{table}
\begin{figure}[t]
\includegraphics[width=\textwidth]{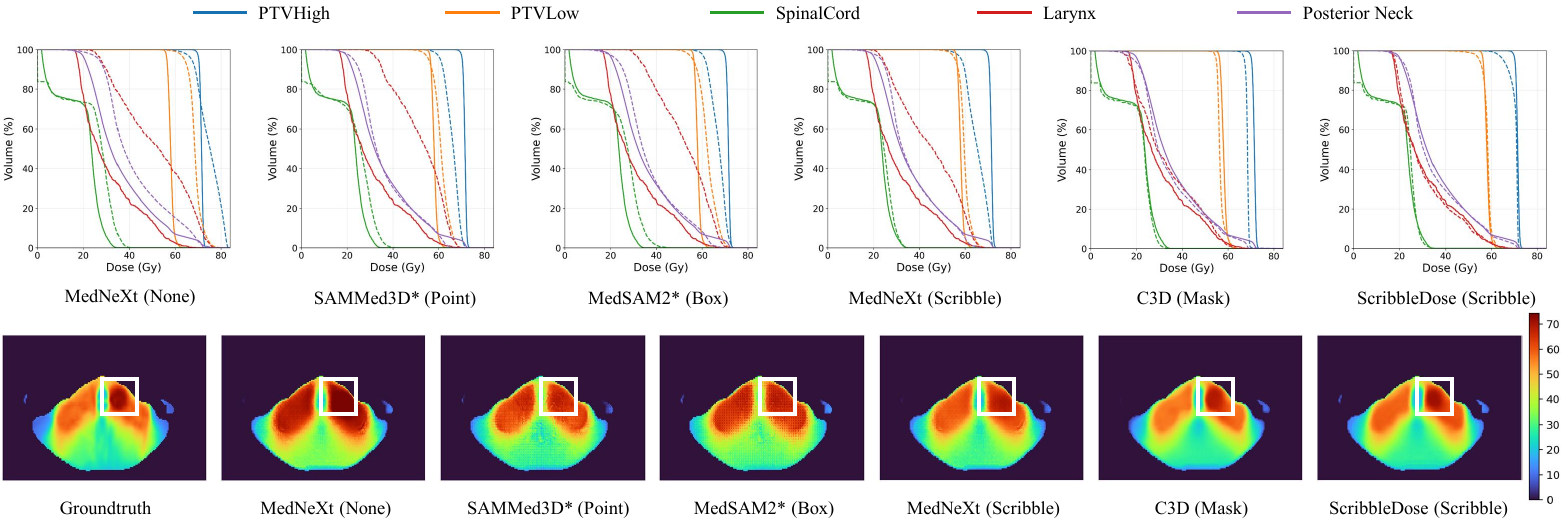}
\caption{Qualitative comparison of predicted dose distributions. \textbf{Top:} DVH of PTVs and OARs. 
Solid lines represent the reference dose distributions, while dashed lines denote the predicted dose distributions.
\textbf{Bottom:} The axial dose maps.} \label{fig3}
\end{figure}

\subsubsection{Ablation Study:}
Table~\ref{tab2} evaluates the contributions of Semantic Guidance (SG) and Morphological Guidance (MG). SG corresponds to the semantic propagation mechanism in the Scribble Completion Module, while MG enforces supervoxel-based boundary constraints for anatomical plausibility. 
Introducing SG substantially reduces the Dose and DVH scores, indicating that semantic propagation effectively reconstructs structure–dose relationships from sparse scribbles. Incorporating MG further improves geometric consistency. The combination of SG and MG achieves the best overall performance, demonstrating that semantic and morphological guidance are complementary for accurate scribble-guided dose prediction.

\subsubsection{Visualization: }Qualitative results in Fig.~\ref{fig3} further demonstrate the effectiveness of the proposed scribble-guided framework. 
DVH comparisons show improved target conformity and competitive OAR sparing, suggesting that sparse scribble guidance effectively approximates dense mask constraints.
The axial dose maps show more concentrated and anatomically consistent high-dose regions (highlighted by the white box), suggesting that sparse scribble supervision can effectively recover meaningful geometric constraints for dose prediction.

\section{Conclusion}
In this work, we propose ScribbleDose, a unified scribble-guided dose prediction framework that alleviates the dependence on fully delineated structural masks.
By introducing the Scribble Completion Module and the Structure-Guided Dose Generation Module, the proposed method establishes a continuous scribble–mask–dose optimization pipeline that restores structure–dose coupling under sparse structural input.
Extensive experiments demonstrate that ScribbleDose achieves competitive performance compared with mask-based approaches while substantially reducing annotation burden.
Despite these encouraging results, this study focuses exclusively on scribble guidance.
Future work will explore a unified sparse-prompt framework that integrates multiple forms of lightweight structural inputs, such as points and bounding boxes, to further enhance flexibility and generalizability in structure-aware dose prediction.

%
%
%
%
\bibliographystyle{IEEEtran}
\bibliography{Paper-1119}

@article{hernandez2020plan,
  title={What is plan quality in radiotherapy? The importance of evaluating dose metrics, complexity, and robustness of treatment plans},
  author={Hernandez, Victor and Hansen, Christian R{\o}nn and Widesott, Lamberto and B{\"a}ck, Anna and Canters, Richard and Fusella, Marco and G{\"o}tstedt, Julia and Jurado-Bruggeman, Diego and Mukumoto, Nobutaka and Kaplan, Laura Patricia and others},
  journal={Radiotherapy and Oncology},
  volume={153},
  pages={26--33},
  year={2020},
  publisher={Elsevier}
}

@article{monshouwer2010practical,
  title={A practical approach to assess clinical planning tradeoffs in the design of individualized IMRT treatment plans},
  author={Monshouwer, Ren{\'e} and Hoffmann, Aswin L and Kunze-Busch, Martina and Bussink, Johan and Kaanders, Johannes HAM and Huizenga, Henk},
  journal={Radiotherapy and Oncology},
  volume={97},
  number={3},
  pages={561--566},
  year={2010},
  publisher={Elsevier}
}

@article{xiong2024automatic,
  title={Automatic planning for functional lung avoidance radiotherapy based on function-guided beam angle selection and plan optimization},
  author={Xiong, Tianyu and Zeng, Guangping and Chen, Zhi and Huang, Yu-Hua and Li, Bing and Zhou, Dejun and Liu, Xi and Sheng, Yang and Ren, Ge and Wu, Qingrong Jackie and others},
  journal={Physics in Medicine \& Biology},
  volume={69},
  number={15},
  pages={155007},
  year={2024},
  publisher={IOP Publishing}
}

@article{nguyen2019feasibility,
  title={A feasibility study for predicting optimal radiation therapy dose distributions of prostate cancer patients from patient anatomy using deep learning},
  author={Nguyen, Dan and Long, Troy and Jia, Xun and Lu, Weiguo and Gu, Xuejun and Iqbal, Zohaib and Jiang, Steve},
  journal={Scientific Reports},
  volume={9},
  number={1},
  pages={1076},
  year={2019},
  publisher={Nature Publishing Group UK London}
}

@article{jiao2023transdose,
  title={TransDose: Transformer-based radiotherapy dose prediction from CT images guided by super-pixel-level GCN classification},
  author={Jiao, Zhengyang and Peng, Xingchen and Wang, Yan and Xiao, Jianghong and Nie, Dong and Wu, Xi and Wang, Xin and Zhou, Jiliu and Shen, Dinggang},
  journal={Medical Image Analysis},
  volume={89},
  pages={102902},
  year={2023},
  publisher={Elsevier}
}

@article{zhou2025novel,
  title={A novel Mamba-based 3D Dose prediction model with channel-aware scan for nasopharyngeal carcinoma radiotherapy},
  author={Zhou, P and Peng, Q and Li, Y and Li, C},
  journal={International Journal of Radiation Oncology, Biology, Physics},
  volume={123},
  number={1},
  pages={e134},
  year={2025},
  publisher={Elsevier}
}

@article{zhang2024dosediff,
  title={DoseDiff: Distance-aware Diffusion model for dose prediction in radiotherapy},
  author={Zhang, Yiwen and Li, Chuanpu and Zhong, Liming and Chen, Zeli and Yang, Wei and Wang, Xuetao},
  journal={IEEE Transactions on Medical Imaging},
  volume={43},
  number={10},
  pages={3621--3633},
  year={2024},
  publisher={IEEE}
}

@article{nijkamp2025current,
  title={Current advances in automation in radiotherapy},
  author={Nijkamp, Jasper and Kn{\"a}usl, Barbara and Aznar, Marianne and Georg, Dietmar and Thorwarth, Daniela and Thwaites, David and Muren, Ludvig P and van der Heide, Uulke A},
  journal={Radiotherapy and Oncology},
  volume={205},
  year={2025},
  publisher={Elsevier}
}

@inproceedings{willems2019feasibility,
  title={Feasibility of CT-only 3D dose prediction for VMAT prostate plans using deep learning},
  author={Willems, Siri and Crijns, Wouter and Sterpin, Edmond and Haustermans, Karin and Maes, Frederik},
  booktitle={Workshop on Artificial Intelligence in Radiation Therapy},
  pages={10--17},
  year={2019},
  organization={Springer}
}

@article{zhang2024generalizable,
  title={Generalizable and promptable artificial intelligence model to augment clinical delineation in radiation oncology},
  author={Zhang, Lian and Liu, Zhengliang and Zhang, Lu and Wu, Zihao and Yu, Xiaowei and Holmes, Jason and Feng, Hongying and Dai, Haixing and Li, Xiang and Li, Quanzheng},
  journal={Medical Physics},
  volume={51},
  number={3},
  pages={2187--2199},
  year={2024},
  publisher={Wiley Online Library}
}

@article{zhang2023segment,
  title={Segment Anything Model (SAM) for radiation oncology},
  author={Zhang, Lian and Liu, Zhengliang and Zhang, Lu and Wu, Zihao and Yu, Xiaowei and Holmes, Jason and Feng, Hongying and Dai, Haixing and Li, Xiang and Li, Quanzheng and others},
  journal={arXiv preprint arXiv:2306.11730},
  year={2023}
}

@article{wen2023transformer,
  title={A Transformer-embedded multi-task model for dose distribution prediction},
  author={Wen, Lu and Xiao, Jianghong and Tan, Shuai and Wu, Xi and Zhou, Jiliu and Peng, Xingchen and Wang, Yan},
  journal={International Journal of Neural Systems},
  volume={33},
  number={08},
  pages={2350043},
  year={2023},
  publisher={World Scientific}
}

@inproceedings{lin2016scribblesup,
  title={ScribbleSup: Scribble-supervised convolutional networks for semantic segmentation},
  author={Lin, Di and Dai, Jifeng and Jia, Jiaya and He, Kaiming and Sun, Jian},
  booktitle={Proceedings of the IEEE Conference on Computer Vision and Pattern Recognition},
  pages={3159--3167},
  year={2016}
}

@inproceedings{zhang2022cyclemix,
  title={CycleMix: A holistic strategy for medical image segmentation from scribble supervision},
  author={Zhang, Ke and Zhuang, Xiahai},
  booktitle={Proceedings of the IEEE/CVF Conference on Computer Vision and Pattern Recognition},
  pages={11656--11665},
  year={2022}
}

@inproceedings{fan2024rmt,
  title={RMT: Retentive networks meet vision Transformers},
  author={Fan, Qihang and Huang, Huaibo and Chen, Mingrui and Liu, Hongmin and He, Ran},
  booktitle={Proceedings of the IEEE/CVF Conference on Computer Vision and Pattern Recognition},
  pages={5641--5651},
  year={2024}
}

@article{gao2025automating,
  title={Automating high quality RT planning at scale},
  author={Gao, Riqiang and Diallo, Mamadou and Liu, Han and Magliari, Anthony and Sackett, Jonathan and Verbakel, Wilko and Meyers, Sandra and Zarepisheh, Masoud and Mcbeth, Rafe and Arberet, Simon and others},
  journal={arXiv e-prints},
  pages={arXiv--2501},
  year={2025}
}

@inproceedings{boettcher2024scribbles,
  title={Scribbles for all: Benchmarking scribble supervised segmentation across datasets},
  author={Boettcher, Wolfgang and Hoyer, Lukas and Unal, Ozan and Lenssen, Jan Eric and Schiele, Bernt},
  booktitle={Advances in Neural Information Processing Systems},
  volume={37},
  pages={46002--46024},
  year={2024}
}

@inproceedings{roy2023mednext,
  title={MedNeXt: Transformer-driven scaling of convnets for medical image segmentation},
  author={Roy, Saikat and Koehler, Gregor and Ulrich, Constantin and Baumgartner, Michael and Petersen, Jens and Isensee, Fabian and Jaeger, Paul F and Maier-Hein, Klaus H},
  booktitle={International Conference on Medical Image Computing and Computer-Assisted Intervention},
  pages={405--415},
  year={2023},
  organization={Springer}
}

@article{MedSAM2,
    title={MedSAM2: Segment anything in 3D medical images and videos},
    author={Ma, Jun and Yang, Zongxin and Kim, Sumin and Chen, Bihui and Baharoon, Mohammed and Fallahpour, Adibvafa and Asakereh, Reza and Lyu, Hongwei and Wang, Bo},
    journal={arXiv preprint arXiv:2504.03600},
    year={2025}
}

@article{wang2025sam,
  title={SAM-Med3D: A vision foundation model for general-purpose segmentation on volumetric medical images},
  author={Wang, Haoyu and Guo, Sizheng and Ye, Jin and Deng, Zhongying and Cheng, Junlong and Li, Tianbin and Chen, Jianpin and Su, Yanzhou and Huang, Ziyan and Shen, Yiqing and others},
  journal={IEEE Transactions on Neural Networks and Learning Systems},
  year={2025},
  publisher={IEEE}
}

@article{liu2021cascade,
  title={A cascade 3D U-Net for dose prediction in radiotherapy},
  author={Liu, Shuolin and Zhang, Jingjing and Li, Teng and Yan, Hui and Liu, Jianfei},
  journal={Medical Physics},
  volume={48},
  number={9},
  pages={5574--5582},
  year={2021},
  publisher={Wiley Online Library}
}

@article{xiong2026generalizable,
  title={A generalizable dose prediction model for automatic radiotherapy planning based on physics-informed priors and large-kernel convolutions},
  author={Xiong, Tianyu and Ren, Ge and Chen, Zhi and Huang, Yu-Hua and Ma, Zongrui and Li, Zihan and Sheng, Yang and Wu, Qingrong Jackie and Cai, Jing},
  journal={Medical Physics},
  volume={53},
  number={1},
  pages={e70272},
  year={2026},
  publisher={Wiley Online Library}
}

@article{hu2023trdosepred,
  title={TrDosePred: A deep learning dose prediction algorithm based on transformers for head and neck cancer radiotherapy},
  author={Hu, Chenchen and Wang, Haiyun and Zhang, Wenyi and Xie, Yaoqin and Jiao, Ling and Cui, Songye},
  journal={Journal of Applied Clinical Medical Physics},
  volume={24},
  number={7},
  pages={e13942},
  year={2023},
  publisher={Wiley Online Library}
}
\end{document}